\documentclass[10pt,twocolumn,letterpaper]{article}

\usepackage{wacv}
\usepackage{times}
\usepackage{epsfig}
\usepackage{graphicx}
\usepackage{amsmath}
\usepackage{amssymb}
\usepackage{algorithm}
\usepackage{algpseudocode}
\usepackage{booktabs}
\usepackage{multirow}
\usepackage{subfigure}

\DeclareMathOperator*{\argmin}{\arg\!\min}

\usepackage{url}

\wacvfinalcopy 

\def\wacvPaperID{165} 

\ifwacvfinal\fi 
\begin{document}

\title{Unconstrained Face Verification using Deep CNN Features}

\author{Jun-Cheng Chen$^1$, Vishal M. Patel$^2$, and Rama Chellappa$^1$\\
1. University of Maryland, College Park\\
2. Rutgers, The State University of New Jersey\\
{\tt\small pullpull@cs.umd.edu, vishal.m.patel@rutgers.edu,
rama@umiacs.umd.edu} }

\maketitle \ifwacvfinal\thispagestyle{empty}\fi

\begin{abstract}


In this paper, we present an algorithm for unconstrained face
verification based on deep convolutional features and evaluate it on
the newly released IARPA Janus Benchmark A (IJB-A) dataset as well
as on the traditional Labeled Face in the Wild (LFW) dataset. The
IJB-A dataset includes real-world unconstrained faces from 500
subjects with full pose and illumination variations which are much
harder than the LFW and Youtube Face (YTF) datasets.  The deep
convolutional neural network (DCNN) is trained using the
CASIA-WebFace dataset. Results of experimental evaluations on the
IJB-A and the LFW datasets are provided.
\end{abstract}

\section{Introduction}

Face verification is one of the core problems in computer vision and
has been actively researched for over two decades
\cite{zhao_face_2003}. In face verification, given two videos or
images, the objective is to determine whether they belong to the
same person.  Many algorithms have been shown to work well on images
that are collected in controlled settings. However, the performance
of these algorithms often degrades significantly on images that have
large variations in pose, illumination, expression, aging,
cosmetics, and occlusion.

To deal with this problem, many methods have focused on learning
invariant and discriminative representation from face images and
videos. One approach is to extract over-complete and
high-dimensional feature representation followed by a learned metric
to project the feature vector into a low-dimensional space and to
compute the similarity score. For instance, the high-dimensional
multi-scale Local Binary Pattern (LBP)\cite{chen_bayesian_2012}
features extracted from local patches around facial landmarks is
reasonably effective for face recognition. Face representation based
on Fisher vector (FV) has also shown to be effective for face
recognition problems
\cite{simonyan_fisher_2013}\cite{parkhi_fisher_2014},
\cite{Chan_FV_BTAS_2015}. However, deep convolutional neural
networks (DCNN) have demonstrated impressive performances on
different tasks such as object recognition
\cite{krizhevsky_imagenet_2012}\cite{szegedy_going_2014}, object
detection \cite{girshick_rich_2014}, and face verification
\cite{schroff_facenet_2015}. It has been shown that a DCNN model can
not only characterize large data variations but also learn a compact
and discriminative feature representation when the size of the
training data is sufficiently large. Once the model is learned, it
is possible to generalize it to other tasks by fine-tuning the
learned model on target datasets \cite{donahue_decaf_2013}. In this
work, we train a DCNN model using a relatively small face dataset,
the CASIA-WebFace \cite{yi_learning_2014}, and compare the
performance of our method with other commercial off-the-shelf face
matchers on the challenging IJB-A dataset which contains significant
variations in pose, illumination, expression, resolution and
occlusion. We also evaluate the performance of the proposed method
on the LFW dataset.

The rest of the paper is organized as follows.  We briefly review
some related works in Section \ref{rel_wor}. Details of the
different components of the proposed method including the DCNN
representation and joint Bayesian metric learning are given in
Section \ref{pro_app}. The protocol and the experimental results are
presented in Section \ref{exp_res}.  Finally, we conclude the paper
in Section~\ref{sec:conc} with a brief summary and discussion.

\begin{figure*}[tb]
\begin{center}
 \includegraphics[width=5.0in]{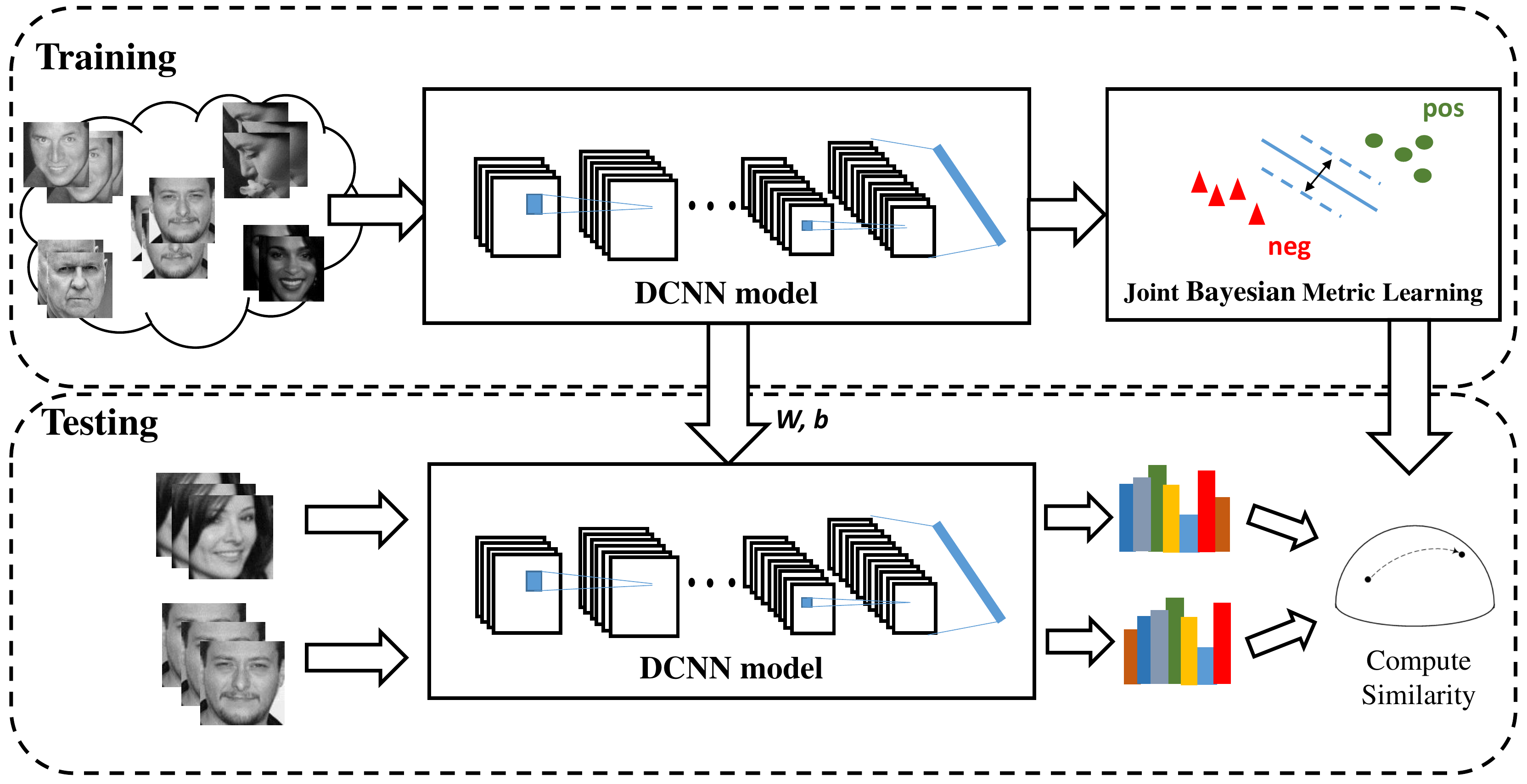}
\end{center}
  \caption{An overview of the proposed DCNN approach for face verification.}
  \label{exp:system_overview}
\end{figure*}

\section{Related Work}
\label{rel_wor} In this section, we briefly review several recent
related works on face verification.

\subsection{Feature Learning}

Learning invariant and discriminative feature representation is the
first step for a face verification system. It can be broadly divided
into two categories: (1) hand-crafted features, and (2) feature
representation learned from data. In the first category, Ahonen
\emph{et al.} \cite{ahonen_face_2006} showed that the Local Binary
Pattern (LBP) is effective for face recognition. Gabor wavelets
\cite{zhang_histogram_2007}\cite{xie_fusing_2010} have also been
widely used to encode multi-scale and multi-orientation information
for face images. Chen \emph{et al.} \cite{chen_blessing_2012}
demonstrated good results for face verification using the
high-dimensional multi-scale LBP features extracted from patches
around facial landmarks. In the second category, Patel \emph{et.
al.} \cite{patel_dictionary_based_2012} and Chen \emph{et. al.}
\cite{chen_dictionary_based_2012}\cite{yc_arfr_2014} applied
dictionary-based approaches for image and video-based face
recognition by learning representative atoms from the data which are
compact and robust to pose and illumination variations .
\cite{simonyan_fisher_2013}\cite{parkhi_fisher_2014}\cite{jc_lffv_2015}
used the FV encoding to generate over-complete and high-dimensional
feature representation for still and video-based face recognition.
Lu \emph{et al.}\cite{lu_joint_2015} proposed a dictionary learning
framework in which the sparse codes of local patches generated from
local patch dictionaries are pooled to generate a high-dimensional
feature vector. The high-dimensionality of feature vectors makes
these methods hard to train and scale to large datasets. However,
advances in deep learning methods have shown that compact and
discriminative representation can be learned using DCNN from very
large datasets. Taigman \emph{et al.} \cite{taigman_deepface_2014}
learned a DCNN model on the frontalized faces generated with a
general 3D shape model from a large-scale face dataset and achieved
better performance than many traditional face verification methods.
Sun \emph{et al.} \cite{sun_deep_2014}\cite{sun_deeply_2014}
achieved results that surpass human performance for face
verification on the LFW dataset using an ensemble of 25 simple DCNN
with fewer layers trained on weakly aligned face images from a much
smaller dataset than the former. Schroff \emph{et al.}
\cite{schroff_facenet_2015} adapted the state-of-the-art deep
architecture for object recognition to face recognition and trained
it on a large-scale unaligned private face dataset with the triplet
loss. This method also achieved top performances on face
verification problems. These works essentially demonstrate the
effectiveness of the DCNN model for feature learning and
detection/recognition/verification problems.

\subsection{Metric Learning}

Learning a similarity measure from data is the other key component
that can boost the performance of a face verification system. Many
approaches have been proposed in the literature that essentially
exploit the label information from face images or face pairs.  For
instance, Weinberger \emph{et al.} \cite{weinberger_distance_2005}
proposed Large Margin Nearest Neighbor(LMNN) metric which enforces
the large margin constraint among all triplets of labeled training
data. Taigman \emph{et al.} \cite{taigman_multiple_2009} learned the
Mahalanobis distance using the Information Theoretic Metric Learning
(ITML) method \cite{davis_information_2007}. Chen \emph{et al.}
\cite{chen_bayesian_2012} proposed a joint Bayesian approach for
face verification which models the joint distribution of a pair of
face images instead of the difference between them, and the ratio of
between-class and within-class probabilities is used as the
similarity measure. Hu \emph{et al.} \cite{hu_discriminative_2014}
learned a discriminative metric within the deep neural network
framework. Huang \emph{et al.} \cite{huang_projection_2015} learned
a projection metric over a set of labeled images which preserves the
underlying manifold structure.

\section{Method}\label{pro_app}
Our approach consists of both training and testing stages. For
training, we first perform face and landmark detection on the
CASIA-WebFace, and the IJB-A datasets to localize and align each
face. Next, we train our DCNN on the CASIA-WebFace and derive the
joint Bayesian metric using the training sets of the IJB-A dataset
and the DCNN features. Then, given a pair of test image sets, we
compute the similarity score based on their DCNN features and the
learned metric. Figure~\ref{exp:system_overview} gives an overview
of our method. The details of each component of our approach are
presented in the following subsections.

\subsection{Preprocessing}
Before training the convolutional network, we perform landmark
detection using the method presented in
\cite{asthana_robust_2013}\cite{asthana_incremental_2014} because of
its ability to be effective on unconstrained faces. Then, each face
is aligned into the canonical coordinate with similarity transform
using the 7 landmark points (\emph{i.e.} two left eye corners, two
right eye corners, nose tip, and two mouth corners). After
alignment, the face image resolution is 100 $\times$ 100 pixels, and
the distance between the centers of two eyes is about 36 pixels.

\subsection{Deep Face Feature Representation}
A DCNN with small filters and very deep architecture (\emph{i.e.} 19
layers in \cite{simonyan_verydeep_2014} and 22 layers in
\cite{szegedy_going_2014}) has shown to produce state-of-the-art
results on many datasets including ImageNet 2014, LFW, and Youtube
Face dataset. Stacking small filters to approximate large filters
and to build very deep convolution networks not only reduces the
number of parameters but also increases the nonlinearity of the
network. In addition, the resulting feature representation is
compact and discriminative.

\begin{figure*}[tb]
\begin{center}
 \includegraphics[width=5.0in]{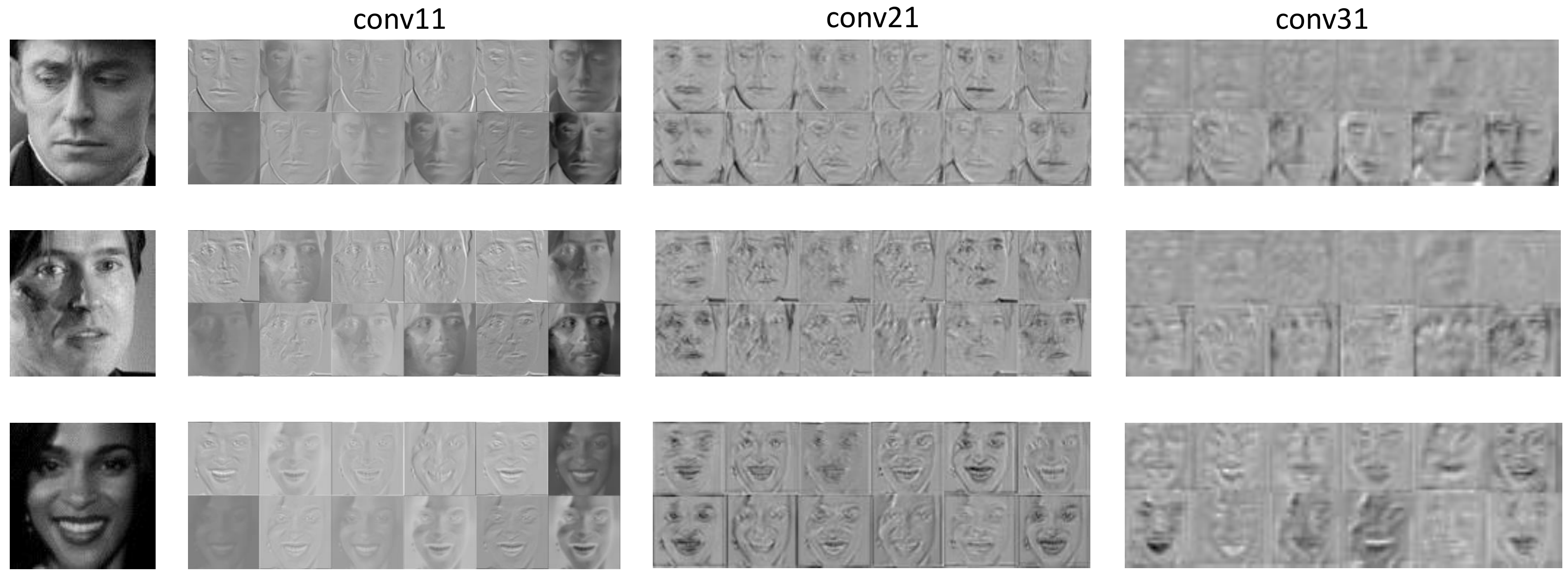}
\end{center}
  \caption{ An illustration of some feature maps of Conv11, Conv21, and Conv31 layers. At the upper layers, the
  feature maps capture more global shape features which are also more robust to illumination changes than Conv11.}
  \label{exp:cnn_features}
\end{figure*}

Our approach is motivated by \cite{yi_learning_2014}. However, we
only consider the identity information per face without modeling the
pair-wise cost. The dimensionality of the input layer is $100 \times
100 \times 1$ for gray-scale images. The network includes 10
convolutional layers, 5 pooling layers and 1 fully connected layer.
The detailed architecture is shown in Table \ref{exp:deep_arch}.
Each convolutional layer is followed by a rectified linear unit
(ReLU) except the last one, Conv52.  Instead of suppressing all the
negative responses to zero using ReLU, we use parametric ReLU
(PReLU)\cite{he_delving_2015}  which allows negative responses that
in turn improves the network performance.  Thus, we use PReLU as an
alternative to ReLU in our work.  Moreover, two local normalization
layers are added after Conv12 and Conv22, respectively to mitigate
the effect of illumination variations. The kernel size of all
filters is $3 \times 3$. The first four pooling layers use the max
operator. To generate a compact and discriminative feature
representation, we use average pooling for the last layer, pool$_5$.
The feature dimensionality of pool$_5$ is thus equal to the number
of channel of Conv52 which is 320. Dropout ratio is set as 0.4 to
regularize Fc6 due to the large number of parameters (\emph{i.e.}
320 $\times$ 10548.). To classify a large number of subjects in the
training data (\emph{i.e.} 10548), this low-dimensional feature
should contain strong discriminative information from all the face
images. Consequently, the pool$_5$ feature is used for face
representation. The extracted features are further $L_2$-normalized
into unit length before the metric learning stage. If there are
multiple frames available for the subject, we use the average of the
pool$_5$ features as the overall feature representation.
Figure~\ref{exp:cnn_features} illustrates some of the extracted
feature maps.

\begin{table*}[tp!]
\centering \small
\begin{tabular}{|c|c|c|c|c|c|}
  \hline
  Name & Type & Filter Size/Stride & Output Size & Depth & $\#$Params\\
  \hline\hline
  Conv11  & convolution & 3$\times$3$\times$1 / 1 & 100$\times$100$\times$32 & 1 & 0.28K\\
  Conv12  & convolution & 3$\times$3$\times$32 / 1 & 100$\times$100$\times$64 & 1 & 18K\\
  \hline
  Pool1   & max pooling & 2$\times$2 / 2 & 50$\times$50$\times$64   & 0 & \\
  \hline
  Conv21  & convolution & 3$\times$3$\times$64 / 1 & 50$\times$50$\times$64   & 1 & 36K\\
  Conv22  & convolution & 3$\times$3$\times$64 / 1 & 50$\times$50$\times$128  & 1 & 72K\\
  \hline
  Pool2   & max pooling & 2$\times$2 / 2 & 25$\times$25$\times$128  & 0 & \\
  \hline
  Conv31  & convolution & 3$\times$3$\times$128 / 1 & 25$\times$25$\times$96   & 1 & 108K\\
  Conv32  & convolution & 3$\times$3$\times$96 / 1 & 25$\times$25$\times$192  & 1 & 162K\\
  \hline
  Pool3   & max pooling & 2$\times$2 / 2 & 13$\times$13$\times$192  & 0 & \\
  \hline
  Conv41  & convolution & 3$\times$3$\times$192 / 1 & 13$\times$13$\times$128  & 1 & 216K\\
  Conv42  & convolution & 3$\times$3$\times$128 / 1 & 13$\times$13$\times$256  & 1 & 288K\\
  \hline
  Pool4   & max pooling & 2$\times$2 / 2 &  7$\times$7$\times$256   & 0 & \\
  \hline
  Conv51  & convolution & 3$\times$3$\times$256 / 1 &  7$\times$7$\times$160   & 1 & 360K\\
  Conv52  & convolution & 3$\times$3$\times$160 / 1 &  7$\times$7$\times$320   & 1 & 450K\\
  \hline
  Pool5   & avg pooling & 7$\times$7 / 1 &  1$\times$1$\times$320   & 0 & \\
  \hline
  Dropout & dropout (40\%) &              &  1$\times$1$\times$320   & 0 & \\
  \hline
  Fc6     & fully connection &              &  10548                   & 1 & 3296K\\
  \hline
  Cost    & softmax          &              &  10548                   & 0 & \\
  \hline\hline
  total   &                  &              &                          & 11 & 5006K \\
  \hline
\end{tabular}
\caption{The architecture of DCNN used in this paper.}
\label{exp:deep_arch}
\end{table*}

\subsection{Joint Bayesian Metric Learning}

To utilize the positive and negative label information available
from the training dataset, we learn a joint Bayesian metric which
has achieved good performances on face verification problems
\cite{chen_bayesian_2012}\cite{cao_practical_2013}. Instead of
modeling the difference vector between two faces, this approach
directly models the joint distribution of feature vectors of both
$i$th and $j$th images, $\{\mathbf{x}_i, \mathbf{x}_j\}$, as a
Gaussian. Let $P(\mathbf{x}_i, \mathbf{x}_j|H_I) \sim N(0,
\boldsymbol{\Sigma}_I)$ when $\mathbf{x}_i$ and $\mathbf{x}_j$
belong to the same class, and $P(\mathbf{x}_i, \mathbf{x}_j|H_E)\sim
N(0, \boldsymbol{\Sigma}_E)$ when they are from different classes.
In addition, each face vector can be modeled as, $\mathbf{x} =
\boldsymbol{\mu} + \boldsymbol{\epsilon}$, where $\boldsymbol{\mu}$
stands for the identity and $\boldsymbol{\epsilon}$ for pose,
illumination, and other variations. Both $\boldsymbol{\mu}$ and
$\boldsymbol{\epsilon}$ are assumed to be independent zero-mean
Gaussian distributions, $N(0, \mathbf{S}_{\mu})$ and $N(0,
\mathbf{S}_{\epsilon})$, respectively.

\noindent The log likelihood ratio of intra- and inter-classes,
$r(\mathbf{x}_i, \mathbf{x}_j)$, can be computed as follows: \small
\begin{equation}
\label{eq:joint_bayesian} r(\mathbf{x}_i, \mathbf{x}_j) = \log
\frac{P(\mathbf{x}_i, \mathbf{x}_j|H_I)}{P(\mathbf{x}_i,
\mathbf{x}_j|H_E)} = \mathbf{x}_i^T\mathbf{M}\mathbf{x}_i +
\mathbf{x}_j^T\mathbf{M}\mathbf{x}_j -
2\mathbf{x}_i^T\mathbf{R}\mathbf{x}_j,
\end{equation}
\normalsize \noindent where $\mathbf{M}$ and $\mathbf{R}$ are both
negative semi-definite matrices. Equation \eqref{eq:joint_bayesian}
can be rewritten as $(\mathbf{x}_i -
\mathbf{x}_j)^T\mathbf{M}(\mathbf{x}_i - \mathbf{x}_j) -
2\mathbf{x}_i^T\mathbf{B}\mathbf{x}_j$ where $\mathbf{B} =
\mathbf{R}-\mathbf{M}$. More details can be found in
\cite{chen_bayesian_2012}. Instead of using the EM algorithm to
estimate $\mathbf{S}_{\mu}$ and $\mathbf{S}_{\epsilon}$, we optimize
the distance in a large-margin framework as follows: \small
\begin{equation}
\displaystyle \argmin_{\mathbf{M}, \mathbf{B}, b}\sum_{i, j} max[1 -
y_{ij}(b -
(\mathbf{x}_i-\mathbf{x}_j)^T\mathbf{M}(\mathbf{x}_i-\mathbf{x}_j) +
2\mathbf{x}_i^T\mathbf{B}\mathbf{x}_j), 0],\\
\label{eq:joint_bayesian_metric}
\end{equation}
\normalsize \noindent where $b \in \mathbb{R}$ is the threshold, and
$y_{ij}$ is the label of a pair: $y_{ij} = 1$ if person $i$ and $j$
are the same and $y_{ij} = -1$, otherwise. For simplicity, we denote
$(\mathbf{x}_i-\mathbf{x}_j)^T\mathbf{M}(\mathbf{x}_i-\mathbf{x}_j)
- 2\mathbf{x}_i^T\mathbf{B}\mathbf{x}_j$ as
$d_{\mathbf{M},\mathbf{B}}(\mathbf{x}_i, \mathbf{x}_j)$.
$\mathbf{M}$ and $\mathbf{B}$ are updated using stochastic gradient
descent as follows and are equally trained on positive and negative
pairs in turn: \small
\begin{equation}
\begin{array}{rl}
\mathbf{M}_{t+1}&=\left\{ \begin{array}{ll}
\mathbf{M}_t\mbox{,} &\mbox{if }y_{ij}(b_t - d_{\mathbf{M},\mathbf{B}}(\mathbf{x}_i, \mathbf{x}_j)) > 1\\
\mathbf{M}_t - \gamma y_{ij}\boldsymbol{\Gamma}_{ij}\mbox{,}
&\mbox{otherwise,} \end{array}\right. \\
\\
\mathbf{B}_{t+1}&=\left\{ \begin{array}{ll}
\mathbf{B}_t\mbox{,} & \mbox{if }y_{ij}(b_t - d_{\mathbf{M},\mathbf{B}}(\mathbf{x}_i, \mathbf{x}_j)) > 1\\
\mathbf{B}_t + 2\gamma y_{ij}\mathbf{x}_i\mathbf{x}_j^T\mbox{,}
&\mbox{otherwise,}
\end{array}\right. \\
\\
b_{t+1}&=\left\{ \begin{array}{ll}
b_t\mbox{,} & \mbox{if }y_{ij}(b_t - d_{\mathbf{M},\mathbf{B}}(\mathbf{x}_i, \mathbf{x}_j)) > 1\\
b_t + \gamma_{b} y_{ij}\mbox{,} &\mbox{otherwise,}
\end{array}\right.
\end{array}
\end{equation}
\normalsize \noindent where
$\boldsymbol{\Gamma}_{ij}=(\mathbf{x}_i-\mathbf{x}_j)(\mathbf{x}_i-\mathbf{x}_j)^T$
and $\gamma$ is the learning rate for $\mathbf{M}$ and $\mathbf{B}$,
and $\gamma_{b}$ for the bias $b$. We use random semi-definite
matrices to initialize both $\mathbf{M} = \mathbf{V}\mathbf{V}^T$
and $\mathbf{B} = \mathbf{W}\mathbf{W}^T$ where both $\mathbf{V}$
and $\mathbf{W} \in \mathbb{R}^{d \times d}$, and $v_{ij}$ and
$w_{ij} \sim N(0,1)$. Note that $\mathbf{M}$ and $\mathbf{B}$ are
updated only when the constraints are violated. In our
implementation, the ratio of the positive and negative pairs that we
generate based on the identity information of the training set is
1:20. In addition, the other reason to train the metric instead of
using traditional EM is that for IJB-A training and test data, some
templates only contain a single image. More details about the IJB-A
dataset are given in Section ~\ref{exp_res}.

\subsection{DCNN Training Details}

The DCNN is implemented using caffe\cite{jia_caffe_2014} and trained
on the CASIA-WebFace dataset. The CASIA-WebFace dataset contains
494,414 face images of 10,575 subjects downloaded from the IMDB
website. After removing the 27 overlapping subjects with the IJB-A
dataset, there are 10548 subjects \footnote{The list of overlapping
subjects is available at
\url{http://www.umiacs.umd.edu/~pullpull/janus_overlap.xlsx}} and
490,356 face images. For each subject, there still exists several
false images with wrong identity labels and few duplicate images.
All images are scaled into $[0,1]$ and subtracted from the mean. The
data is augmented with horizontal flipped face images. We use the
standard batch size 128 for the training phase. Because it only
contains sparse positive and negative pairs per batch in addition to
the false image problems, we do not take the verification cost into
consideration as is done in \cite{sun_deeply_2014}. The initial
negative slope for PReLU is set to 0.25 as suggested in
\cite{he_delving_2015}. The weight decay of all convolutional layers
are set to 0, and the weight decay of the final fully connected
layer to 5e-4. In addition, the learning rate is set to 1e-2
initially and reduced by half every 100,000 iterations. The momentum
is set to 0.9. Finally, we use the snapshot of 1,000,000th iteration
for all our experiments.

\section{Experiments}\label{exp_res}
\label{exp_res} In this section, we present the results of the
proposed approach on the challenging IARPA Janus Benchmark A (IJB-A)
~\cite{klare_janus_2015}, its extended version Janus Challenging set
2 (JANUS CS2) dataset and the LFW dataset.  The JANUS CS2 dataset
contains not only the sampled frames and images in the IJB-A but
also the original videos. The JANUS CS2 dataset\footnote{The JANUS
CS2 dataset is not publicly available yet.} includes much more test
data for identification and verification problems in the defined
protocols than the IJB-A dataset. The receiver operating
characteristic curves (ROC) and the cumulative match characteristic
(CMC) scores are used to evaluate the performance of different
algorithms. The ROC curve measures the performance in the
verification scenarios, and the CMC score measures the accuracy in a
closed set identification scenarios.

\begin{table*}[tb]
\centering \small
\begin{tabular}{|c|ccccc|}
\hline
Probe Template & Rank-1 & Rank-2 & Rank-3 & Rank-4 & Rank-5\\
\hline\hline
  \#Image: 22 & \textcolor{red}{\#Image: 14} & \#Image: 3 & \#Image: 34 & \#Image: 32 & \#Image: 50\\
  \includegraphics[width=0.6in]{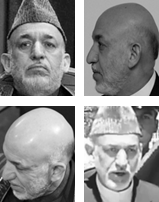}          &  \includegraphics[width=0.6in]{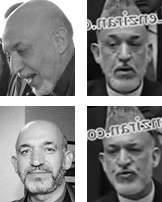}      &   \includegraphics[width=0.6in]{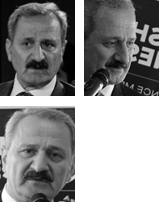}      &    \includegraphics[width=0.6in]{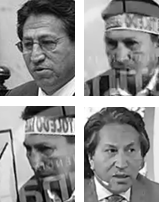}     &   \includegraphics[width=0.6in]{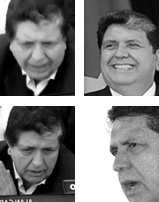}      &   \includegraphics[width=0.6in]{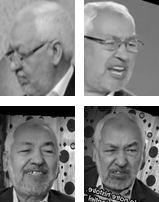}     \\
  Template ID: 2047     &  \textcolor{red}{Template ID: 2030}      &  Template ID: 5794      &  Template ID: 226      & Template ID: 187       &  Template ID: 4726   \\
  Subject ID: 543       &  \textcolor{red}{Subject ID: 543}        &  Subject ID:: 791       &  Subject ID: 102      & Subject ID: 101        &  Subject ID: 404     \\
\hline
  \#Image: 1 & \textcolor{red}{\#Image: 22} & \#Image: 9 & \#Image: 6 & \#Image: 4 & \#Image: 4\\
  \includegraphics[width=0.6in]{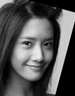}              &    \includegraphics[width=0.6in]{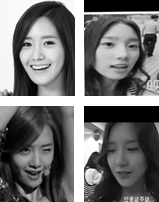}      &   \includegraphics[width=0.6in]{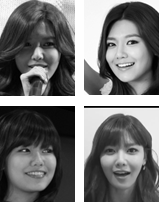}      &    \includegraphics[width=0.6in]{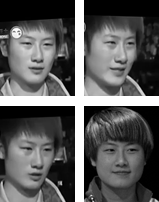}     &   \includegraphics[width=0.6in]{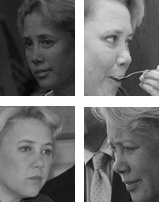}      &   \includegraphics[width=0.6in]{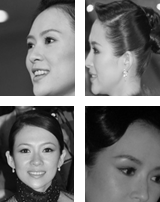}     \\
  Template ID: 2993     &  \textcolor{red}{Template ID: 2992}      &  Template ID: 948  &  Template ID: 1312      & Template ID: 3779       &  Template ID: 5812     \\
  Subject ID: 1559      &  \textcolor{red}{Subject ID: 1559}       &  Subject ID:: 1558 &  Subject ID:: 1704      & Subject ID: 1876       &  Subject ID: 2166     \\
\hline
 \#Image: 1 & \#Image: 25 & \#Image: 7 & \#Image: 3 & \#Image: 32 & \#Image: 6\\
  \includegraphics[width=0.6in]{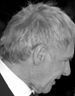}             &    \includegraphics[width=0.6in]{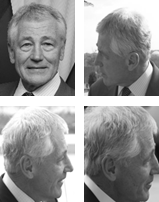}      &   \includegraphics[width=0.6in]{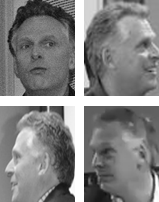}      &    \includegraphics[width=0.6in]{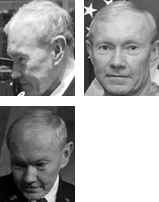}     &   \includegraphics[width=0.6in]{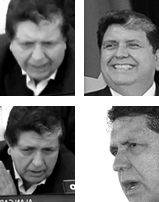}      &   \includegraphics[width=0.6in]{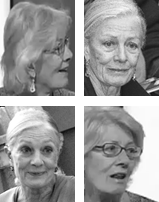}     \\
 Template ID: 2062     &  Template ID: 986      &  Template ID: 5295  &  Template ID: 3729    & Template ID: 187      &  Template ID: 5494     \\
  Subject ID: 158      &  Subject ID: 347       &  Subject ID:: 2058  &  Subject ID: 606      & Subject ID: 101       &  Subject ID: 2102     \\
\hline
\end{tabular}
\vspace{+1mm} \caption{Query results.  The first column shows the
query images from probe templates. The remaining 5 columns show the
corresponding top-5 queried gallery
templates.}\label{exp:query_example}
\end{table*}

\subsection{JANUS-CS2 and IJB-A}



\begin{figure}[htp!]
\begin{center}
 \includegraphics[width=2.3in]{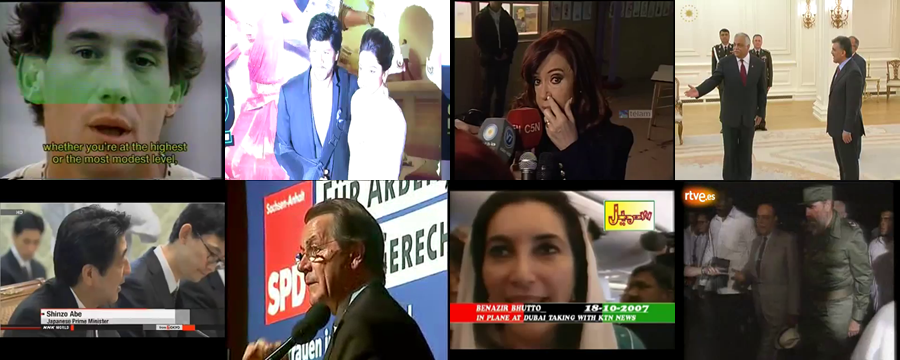}
\end{center}
  \caption{Sample images and frames from the IJB-A and JANUS CS2 datasets. A variety of challenging variations on
  pose, illumination, resolution, occlusion, and image quality are present in these images. }
  \label{exp:sample_janus}
\end{figure}

Both the IJB-A and JANUS CS2 contain 500 subjects with 5,397 images
and 2,042 videos split into 20,412 frames, 11.4 images and 4.2
videos per subject. Sample images and video frames from the datasets
are shown in Fig.~\ref{exp:sample_janus}. The videos are only
released for the JANUS CS2 dataset. The IJB-A evaluation protocol
consists of  verification (1:1 matching) over 10 splits. Each split
contains around 11,748 pairs of templates (1,756 positive and 9,992
negative pairs) on average. Similarly, the identification (1:N
search) protocol also consists of 10 splits which evaluates the
search performance. In each search split, there are about 112
gallery templates and 1763 probe templates (\emph{i.e.} 1,187
genuine probe templates and 576 impostor probe templates). On the
other hand, for the JANUS CS2, there are about 167 gallery templates
and 1763 probe templates and all of them are used for both
identification and verification. The training set for both dataset
contains 333 subjects, and the test set contains 167 subjects. Ten
random splits of training and testing are provided by each
benchmark, respectively. The main differences between IJB-A and
JANUS CS2 evaluation protocol are (1) IJB-A considers the open-set
identification problem and the JANUS CS2 considers the closed-set
identification and (2) IJB-A considers the more difficult pairs
which are the subsets from the JANUS CS2 dataset.

Both the IJB-A and the JANUS CS2 datasets are divided into training
and test sets.  For the test sets of both benchmarks, the image and
video frames of each subject are randomly split into gallery and
probe sets without any overlapping subjects between them.  Unlike
the LFW and YTF datasets which only use a sparse set of negative
pairs to evaluate the verification performance, the IJB-A and JANUS
CS2 both divide the images/video frames into gallery and probe sets
so that it uses all the available positive and negative pairs for
the evaluation. Also, each gallery and probe set consist of multiple
templates. Each template contains a combination of images or frames
sampled from multiple image sets or videos of a subject. For
example, the size of the similarity matrix for JANUS CS2 split1 is
167 $\times$ 1806 where 167 are for the gallery set and 1806 for the
probe set (\emph{i.e.} the same subject reappears multiple times in
different probe templates). Moreover, some templates contain only
one profile face with challenging pose with low quality image. In
contrast to the LFW and YTF datasets which only include faces
detected by the Viola Jones face detector \cite{viola_robust_2004},
the images in the IJB-A and JANUS CS2 contain extreme pose,
illumination and expression variations.  These factors essentially
make the IJB-A and JANUS CS2 challenging face recognition datasets
\cite{klare_janus_2015}.

\subsection{Evaluation on JANUS-CS2 and IJB-A}
\begin{figure*}[htp!]
\centering \subfigure[]{
\includegraphics[height=1.7in]{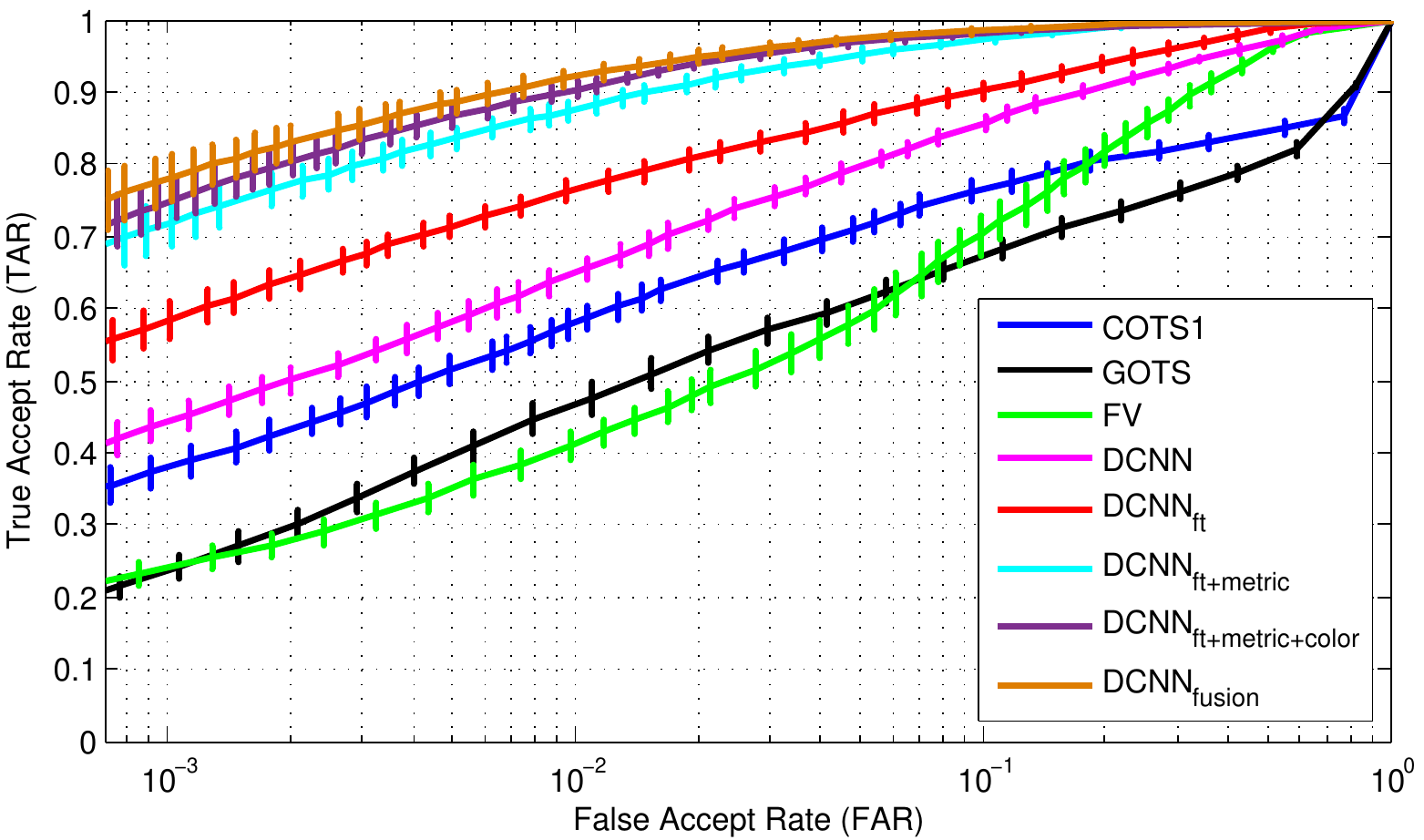}
} \subfigure[]{
\includegraphics[height=1.7in]{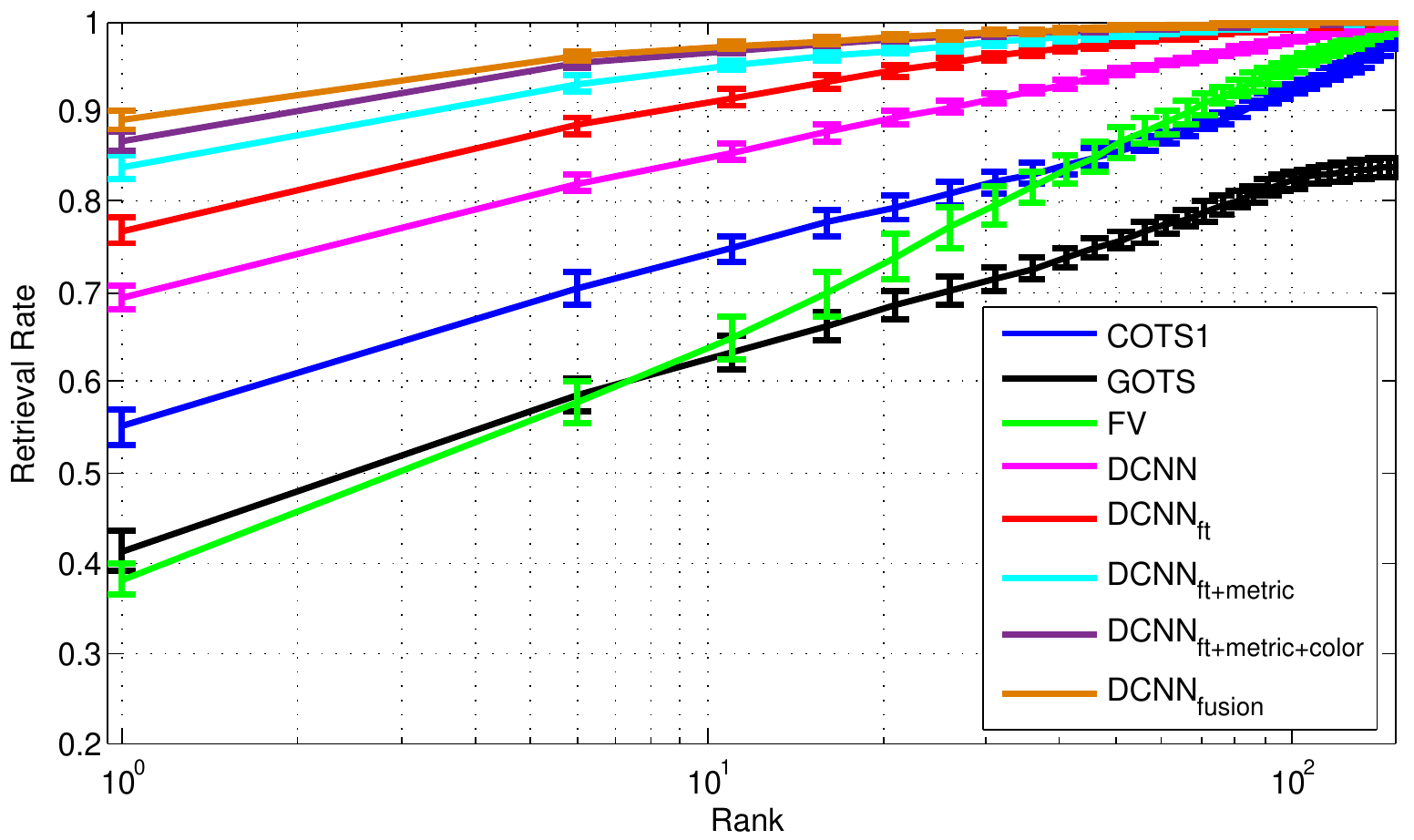}
} \vspace{+1mm}  \caption{Results on the JANUS CS2 dataset.  (a) the
average ROC curves and (b) the average CMC
curves.}\label{exp:janus_roc_cmc}
\end{figure*}

\begin{figure*}[htp!]
\centering \subfigure[]{
\includegraphics[height=1.7in]{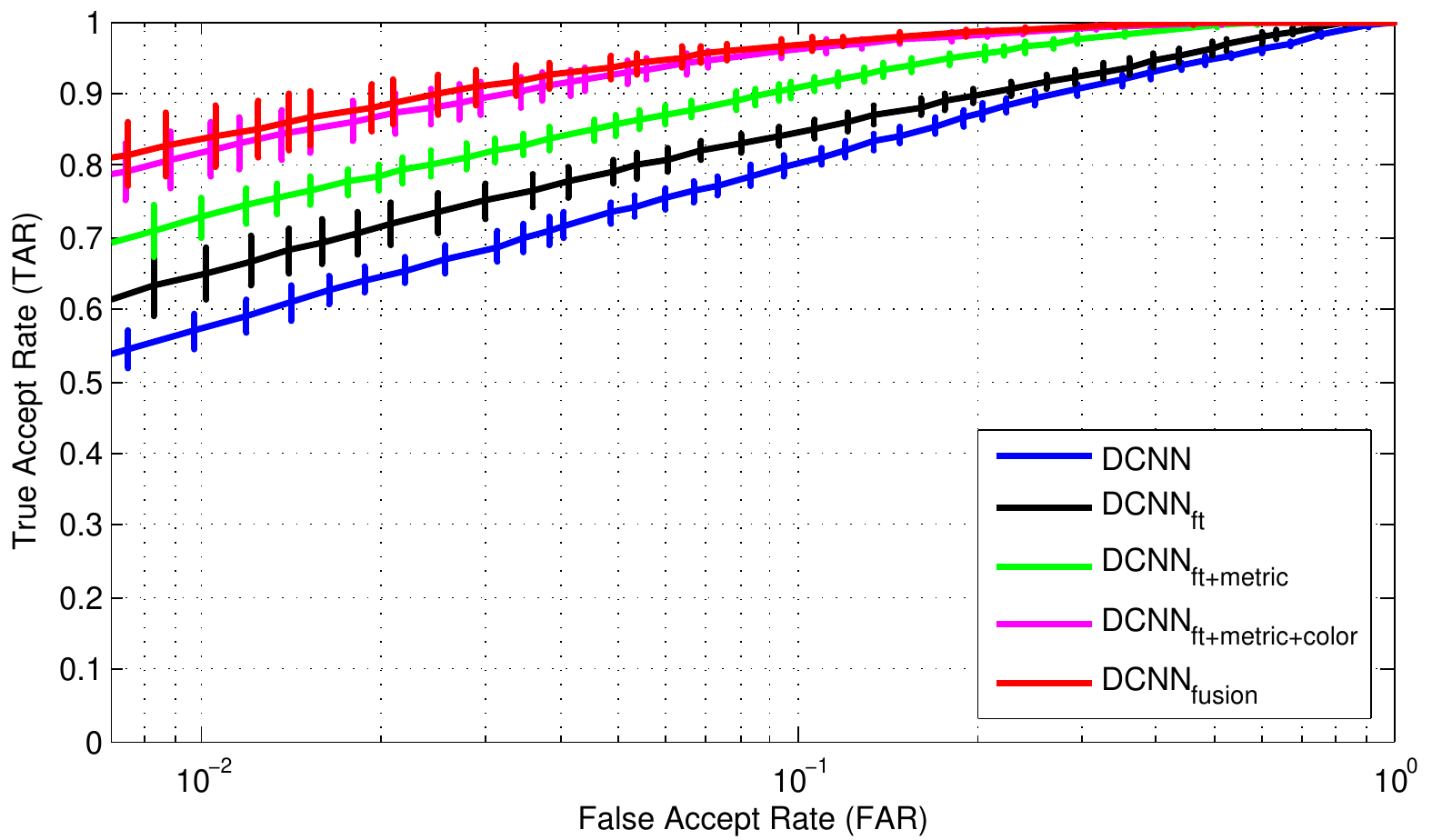}
} \subfigure[]{
\includegraphics[height=1.7in]{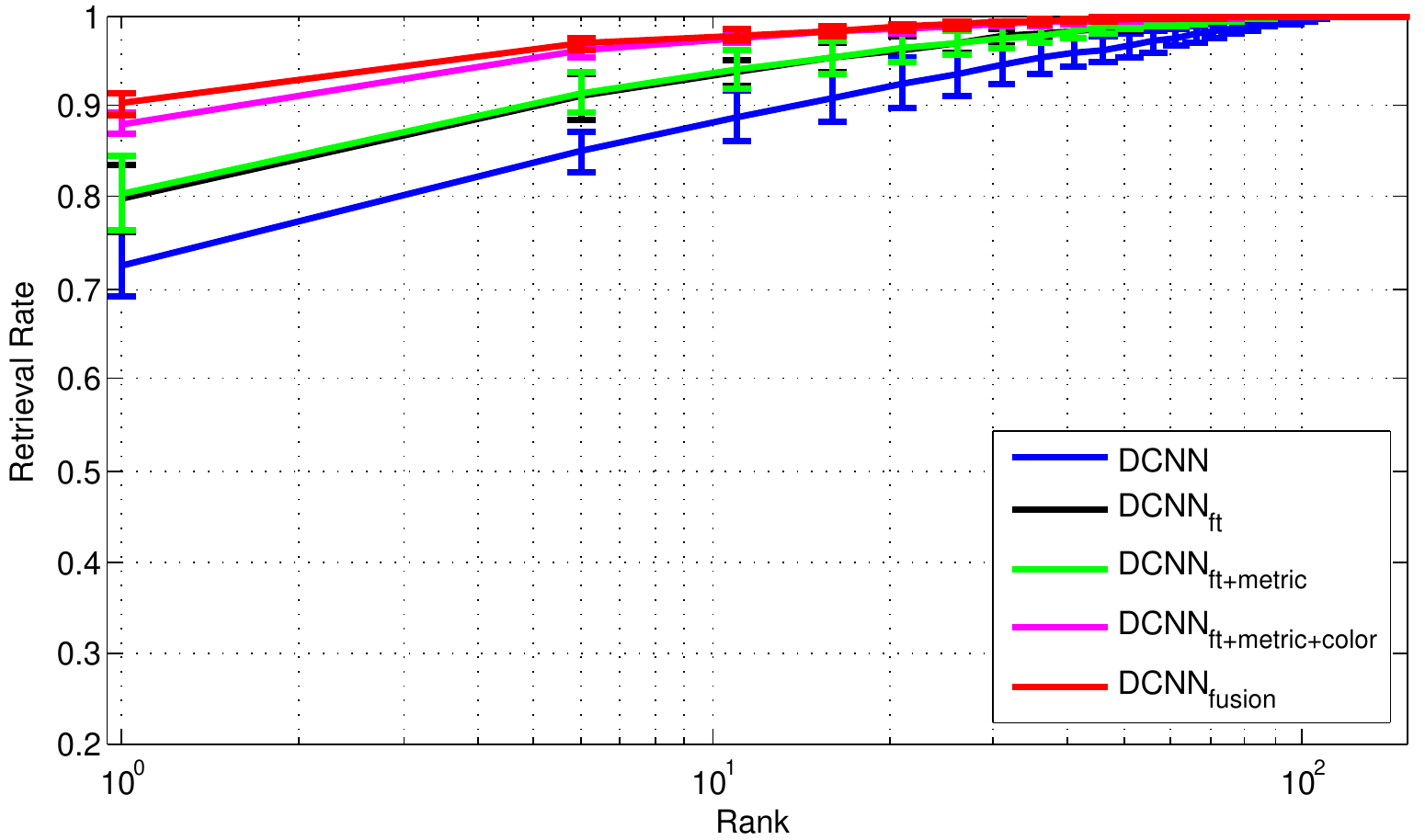}
} \vspace{+1mm} \caption{Results on the IJB-A dataset.  (a) the
average ROC curves for the IJB-A verification protocol and (b) the
average CMC curves for IJB-A identification protocol
 over 10 splits.}
\label{exp:ijba_roc_cmc}
\end{figure*}

For the JANUS CS2 dataset, we compare the results of our DCNN method
with the FV approach proposed in \cite{simonyan_fisher_2013} and two
other commercial off-the-shelf matchers, COTS1 and GOTS
\cite{klare_janus_2015}. The COTS1 and GOTS baselines provided by
JANUS CS2 are the top performers from the most recent NIST FRVT
study \cite{frvt_2014}. The FV method is trained on the LFW dataset
which contains few faces with extreme pose. Therefore, we use the
pose information estimated from the landmark detector and select
face images/video frames whose yaw angle are less than or equal to
$\pm$25 degrees for each gallery and probe set. If there are no
images/frames satisfying the constraint, we choose the one closest
to the frontal one. However, for the DCNN method, we use all the
frames without applying the same selection strategy. \footnote{We
fix the typos in \cite{chen_unconstrained_2015} that the selection
strategy is only applied to FV-based method, not for DCNN.}
Figures~\ref{exp:janus_roc_cmc} and \ref{exp:ijba_roc_cmc} show the
ROC curves and the CMC curves, respectively for the verification
results using the previously described protocol where DCNN means
using DCNN feature with cosine distance, ``ft" means finetuning on
the training data, ``metric" means applying Joint Bayesian metric
learning, and ``color" means to use all of the RGB images instead of
gray-scale images. For the results of DCNN$_{ft+metric}$, besides
finetuning and metric learning, we also replace ReLU with PReLU and
apply data augmentation (\emph{i.e.} randomly cropping 100 $\times$
100-pixel subregions from a 125 $\times$ 125 region). For
DCNN$_{ft+metric+color}$\footnote{DCNN$_{ft+metric+color}$ and
DCNN$_{fusion}$ are our improved results for JANUS CS2 and IJB-A
datasets obtained after the paper was accepted.}, we further use RGB
images and larger face regions. (\emph{i.e.} we use 125 $\times$
125-pixel face regions and resize them into 100 $\times$ 100-pixel
ones.) Then, we show the fusion results, DCNN$_{fusion}$, by
directly summing the similarity scores of two models,
DCNN$_{ft+metric}$ and DCNN$_{ft+metric+color}$, where
DCNN$_{ft+metric}$ is trained on gray-scale images with smaller face
regions and DCNN$_{ft+metric+color}$ is trained on RGB images with
larger face regions. From these figures, we can clearly see the
impact of each component to the improvement of final identification
and verification results. From the ROC and CMC curves, we see that
the DCNN method performs better than other competitive methods. This
can be attributed to the fact that the DCNN model does capture face
variations over a large dataset and generalizes well to a new small
dataset.

We illustrate the query samples in Table ~\ref{exp:query_example}.
The first column shows the query images from the probe templates.
The remaining five columns show the corresponding top-5 queried
gallery templates (\emph{i.e.} rank-1 means the most similar one,
rank-2 the second most similar, etc.). For the first two rows, our
approach can successfully find the subjects in rank 1. For the
third, the query template only contains one image with extreme pose.
However, in the corresponding gallery template for the same subject,
it happens to contain only near-frontal faces. Thus, it failed to
find the subject within the top-5 matches. To solve the pose
generalization problem of CNN features, one possible solution is to
augment the templates by synthesizing faces in various poses with
the help of a generic 3D model. We plan to pursue this approach in
the near future, and we leave it for the future work.

While this paper was under preparation, the authors became aware of
\cite{wang_face_2015}, which also proposes a CNN-based approach for
face verification/identification and evaluates it on the IJB-A
dataset.  The method proposed in \cite{wang_face_2015} combines the
features from seven independent DCNN models.  With finetuning on the
JANUS training data and metric learning, our approach works
comparable to \cite{wang_face_2015} as shown in
Figure~\ref{exp:ijba_roc_cmc}. Furthermore, with the replacement of
ReLU with PReLU and data augmentation, our approach significantly
outperforms \cite{wang_face_2015} with only a single model.

\begin{table*}[htp!]
\centering \footnotesize
\begin{tabular}{|c|c|c|c|c|c|c|}
  \hline
  IJB-A-Verif&\cite{wang_face_2015}&DCNN&DCNN$_{ft}$&DCNN$_{ft+m}$&DCNN$_{ft+m+c}$&DCNN$_{fusion}$\\
  \hline
  FAR=1e-2&0.732$\pm$0.033&0.573$\pm0.024$&0.64$\pm$0.045&0.787$\pm$0.043&0.818$\pm$0.037&\textbf{0.838$\pm$0.042}\\
  FAR=1e-1&0.895$\pm$0.013&0.8$\pm$0.012&0.883$\pm$0.012&0.947$\pm$0.011&0.961$\pm$0.01&\textbf{0.967$\pm$0.009}\\
  \hline\hline
  IJB-A-Ident&\cite{wang_face_2015}&DCNN&DCNN$_{ft}$&DCNN$_{ft+m}$\footnote{}&DCNN$_{ft+m+c}$&DCNN$_{fusion}$\\
  \hline
  Rank-1 &0.820$\pm$0.024&0.726$\pm$0.034&0.799$\pm$0.036&0.852$\pm$0.018&0.882$\pm$0.01&\textbf{0.903 $\pm$0.012}\\
  Rank-5 &0.929$\pm$0.013&0.84$\pm$0.023&0.901$\pm$0.025&0.937$\pm$0.01&0.957$\pm$0.07&\textbf{0.965$\pm$0.008}\\
  Rank-10&$N/A$&0.884$\pm$0.025&0.934$\pm$0.016&0.954$\pm$0.007&0.974$\pm$0.005&\textbf{0.977$\pm$0.007}\\
  \hline
\end{tabular}
\vspace{+1mm} \caption{Results on the IJB-A dataset.  The TAR of all
the approaches at FAR=0.1 and 0.01 for the ROC curves. The Rank-1,
Rank-5, and Rank-10 retrieval accuracies of the CMC curves where
subscripts \emph{ft}, \emph{m} and \emph{c} stand for finetuning,
metric, and color respectively.} \label{exp:roc_cmc_scores_ijba}
\end{table*}
\footnotetext[5]{We correct the number reported in
\cite{chen_unconstrained_2015} previously for the IJB-A
identification task because one split of the identification task was
performed partially due to the corrupted metadata. (i.e. Some images
were missing at that time. The current metadata of IJB-A has fixed
those errors already.)}

\begin{table*}[htp!]
\centering \footnotesize
\begin{tabular}{|c|c|c|c|c|c|c|c|c|}
  \hline
  CS2-Verif&COTS1&GOTS&FV\cite{simonyan_fisher_2013}&DCNN&DCNN$_{ft}$&DCNN$_{ft+m}$&DCNN$_{ft+m+c}$&DCNN$_{fusion}$\\
  \hline
  FAR=1e-2&0.581$\pm$0.054&0.467$\pm$0.066&0.411$\pm$0.081&0.649$\pm$0.015&0.765$\pm$0.014&0.876$\pm$0.013&0.904$\pm$0.011&\textbf{0.921$\pm$0.013}\\
  FAR=1e-1&0.767$\pm$0.015&0.675$\pm$0.015&0.704$\pm$0.028&0.855$\pm$0.01&0.902$\pm$0.011&0.973$\pm$0.005&0.983$\pm$0.004&\textbf{0.985$\pm$0.004}\\
  \hline\hline
  CS2-Ident&COTS1&GOTS&FV \cite{simonyan_fisher_2013}&DCNN&DCNN$_{ft}$&DCNN$_{ft+m}$&DCNN$_{ft+m+c}$&DCNN$_{fusion}$\\
  \hline
  Rank-1&0.551$\pm$0.03&0.413$\pm$0.022&0.381$\pm$0.018&0.694$\pm$0.012&0.768$\pm$0.013&0.838$\pm$0.012&0.867$\pm$0.01&\textbf{0.891$\pm$0.01}\\
  Rank-5&0.694$\pm$0.017&0.571$\pm$0.017&0.559$\pm$0.021&0.809$\pm$0.011&0.874$\pm$0.01&0.924$\pm$0.009&0.949$\pm$0.005&\textbf{0.957$\pm$0.007}\\
  Rank-10&0.741$\pm$0.017&0.624$\pm$0.018&0.637$\pm$0.025&0.85$\pm$0.009&0.91$\pm$0.008&0.949$\pm$0.006&0.966$\pm$0.005&\textbf{0.972$\pm$0.005}\\
  \hline
\end{tabular}
\vspace{+1mm} \caption{Results on the JANUS CS2 dataset.  The TAR of
all the approaches at FAR=0.1 and 0.01 for the ROC curves. The
Rank-1, Rank-5, and Rank-10 retrieval accuracies of the CMC curves
where subscripts \emph{ft}, \emph{m} and \emph{c} stand for
finetuning, metric, and color respectively.}
\label{exp:roc_cmc_scores}
\end{table*}

\begin{table*}[htp!]
\centering \small
\begin{tabular}{|l|c|l|l|l|}
  \hline
  Method          &\#Net& Training Set & Metric & Mean Accuracy $\pm$ Std\\
  \hline\hline
  DeepFace \cite{taigman_deepface_2014}&  1  & 4.4 million images of 4,030 subjects, private     & cosine                    & 95.92\% $\pm$ 0.29\% \\
  DeepFace                             &  7  & 4.4 million images of 4,030 subjects, private     & unrestricted, SVM         & 97.35\% $\pm$ 0.25\% \\
  DeepID2 \cite{sun_deeply_2014}       &  1  & 202,595 images of 10,117 subjects, private        & unrestricted, Joint-Bayes & 95.43\% \\
  DeepID2                              & 25  & 202,595 images of 10,117 subjects, private        & unrestricted, Joint-Bayes & 99.15\% $\pm$ 0.15\% \\
  DeepID3 \cite{sun_deepid3_2015}      & 50  & 202,595 images of 10,117 subjects, private        & unrestricted, Joint-Bayes & 99.53\% $\pm$ 0.10\% \\
  FaceNet \cite{schroff_facenet_2015}  &  1  & 260 million images of 8 million subjects, private & L2                        & 99.63\% $\pm$ 0.09\% \\
  Yi et al. \cite{yi_learning_2014}    &  1  & 494,414 images of 10,575 subjects, public         & cosine                    & 96.13\% $\pm$ 0.30\% \\
  Yi et al.                            &  1  & 494,414 images of 10,575 subjects, public         & unrestricted, Joint-Bayes & 97.73\% $\pm$ 0.31\% \\
  Wang et al. \cite{wang_face_2015}    &  1  & 494,414 images of 10,575 subjects, public         & cosine                    & 96.95\% $\pm$ 1.02\% \\
  Wang et al.                          &  7  & 494,414 images of 10,575 subjects, public         & cosine                    & 97.52\% $\pm$ 0.76\% \\
  Wang et al.                          &  1  & 494,414 images of 10,575 subjects, public         & unrestricted, Joint-Bayes & 97.45\% $\pm$ 0.99\% \\
  Wang et al.                          &  7  & 494,414 images of 10,575 subjects, public         & unrestricted, Joint-Bayes & 98.23\% $\pm$ 0.68\% \\
  Human, funneled \cite{wang_face_2015}& N/A & N/A & N/A & 99.20\% \\
  \hline
  Ours            &  1  & 490,356 images of 10,548 subjects, public         & cosine                    & 97.15\% $\pm$ 0.7\% \\
  Ours            &  1  & 490,356 images of 10,548 subjects, public         & unrestricted, Joint-Bayes & 97.45\% $\pm$ 0.7\% \\
  \hline
\end{tabular}
\vspace{+1mm} \caption{Accuracy of different methods on the LFW
dataset.} \label{exp:acc_lfw}
\end{table*}

\subsection{Labeled Face in the Wild}
We also evaluate our approach on the well-known LFW dataset using
the standard protocol which defines 3,000 positive pairs and 3,000
negative pairs in total and further splits them into 10 disjoint
subsets for cross validation. Each subset contains 300 positive and
300 negative pairs. It contains 7,701 images of 4,281 subjects. We
compare the mean accuracy of the proposed deep model with other
state-of-the-art deep learning-based methods: DeepFace
\cite{taigman_deepface_2014}, DeepID2 \cite{sun_deeply_2014},
DeepID3 \cite{sun_deepid3_2015}, FaceNet
\cite{schroff_facenet_2015}, Yi \emph{et al.}
\cite{yi_learning_2014}, Wang \emph{et al.} \cite{wang_face_2015},
and human performance on the ``funneled" LFW images.  The results
are summarized in Table ~\ref{exp:acc_lfw}.  It can be seen from
this table that our approach performs comparably to other deep
learning-based methods.  Note that some of the deep learning-based
methods compared in Table~\ref{exp:acc_lfw} use millions of data
samples for training the model.  Whereas we use only the CASIA
dataset for training our model which has less than 500K images.

\subsection{Run Time}
The DCNN model is trained for about 9 days using NVidia Tesla K40.
The feature extraction time takes about 0.006 second per face image.
In future, the supervised information will be fed into the
intermediate layers to make the model more discriminative and also
to converge faster.

\section{Conclusion}\label{sec:conc}

In this paper, we study the performance of a DCNN method on a newly
released challenging face verification dataset, IARPA Benchmark A,
which contains faces with full pose, illumination, and other
difficult conditions. It was shown that the DCNN approach can learn
a robust model from a large dataset characterized by face variations
and generalizes well to another dataset. Experimental results
demonstrate that the performance of the proposed DCNN on the IJB-A
dataset is much better than the FV-based method and other commercial
off-the-shelf matchers and is competitive for the LFW dataset.

For future work, we plan to directly train a Siamese network using
all the available positive and negative pairs from CASIA-Webface and
IJB-A training datasets to fully utilize the discriminative
information for realizing better performance.

\section{Acknowledgments}
This research is based upon work supported by the Office of the
Director of National Intelligence (ODNI), Intelligence Advanced
Research Projects Activity (IARPA), via IARPA R\&D Contract No.
2014-14071600012. The views and conclusions contained herein are
those of the authors and should not be interpreted as necessarily
representing the official policies or endorsements, either expressed
or implied, of the ODNI, IARPA, or the U.S. Government. The U.S.
Government is authorized to reproduce and distribute reprints for
Governmental purposes notwithstanding any copyright annotation
thereon. We thank NVIDIA for donating of the K40 GPU used in this
work.

{\small
\bibliographystyle{ieee}
\bibliography{refs,refs_vishal,refs_swami}

\begin{thebibliography}{10}\itemsep=-1pt

\bibitem{ahonen_face_2006}
T.~Ahonen, A.~Hadid, and M.~Pietikainen.
\newblock Face description with local binary patterns: Application to face
  recognition.
\newblock {\em {IEEE} Transactions on Pattern Analysis and Machine
  Intelligence}, 28(12):2037--2041, 2006.

\bibitem{asthana_robust_2013}
A.~Asthana, S.~Zafeiriou, S.~Y. Cheng, and M.~Pantic.
\newblock Robust discriminative response map fitting with constrained local
  models.
\newblock In {\em {IEEE} Conference on Computer Vision and Pattern
  Recognition}, pages 3444--3451, 2013.

\bibitem{asthana_incremental_2014}
A.~Asthana, S.~Zafeiriou, S.~Y. Cheng, and M.~Pantic.
\newblock Incremental face alignment in the wild.
\newblock In {\em {IEEE} Conference on Computer Vision and Pattern
  Recognition}, pages 1859--1866, 2014.

\bibitem{cao_practical_2013}
X.~D. Cao, D.~Wipf, F.~Wen, G.~Q. Duan, and J.~Sun.
\newblock A practical transfer learning algorithm for face verification.
\newblock In {\em IEEE International Conference on Computer Vision}, pages
  3208--3215. IEEE, 2013.

\bibitem{chen_bayesian_2012}
D.~Chen, X.~D. Cao, L.~W. Wang, F.~Wen, and J.~Sun.
\newblock Bayesian face revisited: A joint formulation.
\newblock In {\em European Conference on Computer Vision}, pages 566--579.
  2012.

\bibitem{chen_blessing_2012}
D.~Chen, X.~D. Cao, F.~Wen, and J.~Sun.
\newblock Blessing of dimensionality: High-dimensional feature and its
  efficient compression for face verification.
\newblock In {\em {IEEE} Conference on Computer Vision and Pattern
  Recognition}, 2013.

\bibitem{jc_lffv_2015}
J.-C. Chen, V.~M. Patel, and R.~Chellappa.
\newblock Landmark-based fisher vector representation for video-based face
  verification.
\newblock In {\em {IEEE} Conference on Image Processing}, 2015.

\bibitem{chen_unconstrained_2015}
J.-C. Chen, V.~M. Patel, and R.~Chellappa.
\newblock Unconstrained face verification using deep cnn features.
\newblock {\em arXiv preprint arXiv:1508.01722}, 2015.

\bibitem{Chan_FV_BTAS_2015}
J.-C. Chen, S.~Sankaranarayanan, V.~M. Patel, and R.~Chellappa.
\newblock Unconstrained face verification using fisher vectors computed from
  frontalized faces.
\newblock In {\em IEEE International Conference on Biometrics: Theory,
  Applications and Systems}, 2015.

\bibitem{yc_arfr_2014}
Y.-C. Chen, V.~M. Patel, R.~Chellappa, and P.~J. Phillips.
\newblock Adaptive representations for video-based face recognition across
  pose.
\newblock In {\em {IEEE} Winter Conference on Applications of Computer Vision},
  2014.

\bibitem{chen_dictionary_based_2012}
Y.-C. Chen, V.~M. Patel, P.~J. Phillips, and R.~Chellappa.
\newblock Dictionary-based face recognition from video.
\newblock In {\em European Conference on Computer Vision}, pages 766--779.
  2012.

\bibitem{davis_information_2007}
J.~V. Davis, B.~Kulis, P.~Jain, S.~Sra, and I.~S. Dhillon.
\newblock Information-theoretic metric learning.
\newblock In {\em International Conference on Machine learning}, pages
  209--216, 2007.

\bibitem{donahue_decaf_2013}
J.~Donahue, Y.~Jia, O.~Vinyals, J.~Hoffman, N.~Zhang, E.~Tzeng, and T.~Darrell.
\newblock Decaf: A deep convolutional activation feature for generic visual
  recognition.
\newblock {\em arXiv preprint arXiv:1310.1531}, 2013.

\bibitem{girshick_rich_2014}
R.~Girshick, J.~Donahue, T.~Darrell, and J.~Malik.
\newblock Rich feature hierarchies for accurate object detection and semantic
  segmentation.
\newblock In {\em IEEE Conference on Computer Vision and Pattern Recognition},
  pages 580--587, 2014.

\bibitem{frvt_2014}
P.~Grother and M.~Ngan.
\newblock Face recognition vendor test(frvt): Performance of face
  identification algorithms.
\newblock {\em NIST Interagency Report 8009}, 2014.

\bibitem{he_delving_2015}
K.~He, X.~Zhang, S.~Ren, and J.~Sun.
\newblock Delving deep into rectifiers: Surpassing human-level performance on
  imagenet classification.
\newblock {\em arXiv preprint arXiv:1502.01852}, 2015.

\bibitem{hu_discriminative_2014}
J.~Hu, J.~Lu, and Y.-P. Tan.
\newblock Discriminative deep metric learning for face verification in the
  wild.
\newblock In {\em IEEE Conference on Computer Vision and Pattern Recognition},
  pages 1875--1882, 2014.

\bibitem{huang_projection_2015}
Z.~Huang, R.~Wang, S.~Shan, and X.~Chen.
\newblock Projection metric learning on {G}rassmann manifold with application
  to video based face recognition.
\newblock In {\em IEEE Conference on Computer Vision and Pattern Recognition},
  pages 140--149, 2015.

\bibitem{jia_caffe_2014}
Y.~Jia, E.~Shelhamer, J.~Donahue, S.~Karayev, J.~Long, R.~Girshick,
  S.~Guadarrama, and T.~Darrell.
\newblock Caffe: Convolutional architecture for fast feature embedding.
\newblock In {\em ACM International Conference on Multimedia}, pages 675--678,
  2014.

\bibitem{klare_janus_2015}
B.~F. Klare, B.~Klein, E.~Taborsky, A.~Blanton, J.~Cheney, K.~Allen,
  P.~Grother, A.~Mah, M.~Burge, and A.~K. Jain.
\newblock Pushing the frontiers of unconstrained face detection and
  recognition: {IARPA Janus Benchmark A}.
\newblock In {\em IEEE Conference on Computer Vision and Pattern Recognition},
  2015.

\bibitem{krizhevsky_imagenet_2012}
A.~Krizhevsky, I.~Sutskever, and G.~E. Hinton.
\newblock Imagenet classification with deep convolutional neural networks.
\newblock In {\em Advances in Neural Information Processing Systems}, pages
  1097--1105, 2012.

\bibitem{lu_joint_2015}
J.~Lu, V.~E. Liong, G.~Wang, and P.~Moulin.
\newblock Joint feature learning for face recognition.
\newblock {\em IEEE Transactions on Information Forensics and Security},
  PP(99):1--1, 2015.

\bibitem{parkhi_fisher_2014}
O.~M. Parkhi, K.~Simonyan, A.~Vedaldi, and A.~Zisserman.
\newblock A compact and discriminative face track descriptor.
\newblock In {\em IEEE Conference on Computer Vision and Pattern Recognition},
  2014.

\bibitem{patel_dictionary_based_2012}
V.~M. Patel, T.~Wu, S.~Biswas, P.~J. Phillips, and R.~Chellappa.
\newblock Dictionary-based face recognition under variable lighting and pose.
\newblock {\em {IEEE} Transactions on Information Forensics and Security},
  7(3):954--965, 2012.

\bibitem{schroff_facenet_2015}
F.~Schroff, D.~Kalenichenko, and J.~Philbin.
\newblock Facenet: A unified embedding for face recognition and clustering.
\newblock {\em arXiv preprint arXiv:1503.03832}, 2015.

\bibitem{simonyan_fisher_2013}
K.~Simonyan, O.~M. Parkhi, A.~Vedaldi, and A.~Zisserman.
\newblock Fisher vector faces in the wild.
\newblock In {\em British Machine Vision Conference}, volume~1, page~7, 2013.

\bibitem{simonyan_verydeep_2014}
K.~Simonyan and A.~Zisserman.
\newblock Very deep convolutional networks for large-scale image recognition.
\newblock {\em arXiv preprint arXiv:1409.1556}, 2014.

\bibitem{sun_deep_2014}
Y.~Sun, Y.~Chen, X.~Wang, and X.~Tang.
\newblock Deep learning face representation by joint
  identification-verification.
\newblock In {\em Advances in Neural Information Processing Systems}, pages
  1988--1996, 2014.

\bibitem{sun_deepid3_2015}
Y.~Sun, D.~Liang, X.~Wang, and X.~Tang.
\newblock Deepid3: Face recognition with very deep neural networks.
\newblock {\em arXiv preprint arXiv:1502.00873}, 2015.

\bibitem{sun_deeply_2014}
Y.~Sun, X.~Wang, and X.~Tang.
\newblock Deeply learned face representations are sparse, selective, and
  robust.
\newblock {\em arXiv preprint arXiv:1412.1265}, 2014.

\bibitem{szegedy_going_2014}
C.~Szegedy, W.~Liu, Y.~Jia, P.~Sermanet, S.~Reed, D.~Anguelov, D.~Erhan,
  V.~Vanhoucke, and A.~Rabinovich.
\newblock Going deeper with convolutions.
\newblock {\em arXiv preprint arXiv:1409.4842}, 2014.

\bibitem{taigman_multiple_2009}
Y.~Taigman, L.~Wolf, and T.~Hassner.
\newblock Multiple one-shots for utilizing class label information.
\newblock In {\em British Machine Vision Conference}, pages 1--12, 2009.

\bibitem{taigman_deepface_2014}
Y.~Taigman, M.~Yang, M.~A. Ranzato, and L.~Wolf.
\newblock Deepface: Closing the gap to human-level performance in face
  verification.
\newblock In {\em IEEE Conference on Computer Vision and Pattern Recognition},
  pages 1701--1708, 2014.

\bibitem{viola_robust_2004}
P.~Viola and M.~J. Jones.
\newblock Robust real-time face detection.
\newblock {\em International journal of computer vision}, 57(2):137--154, 2004.

\bibitem{wang_face_2015}
D.~Wang, C.~Otto, and A.~K. Jain.
\newblock Face search at scale: 80 million gallery.
\newblock {\em arXiv preprint arXiv:1507.07242}, 2015.

\bibitem{weinberger_distance_2005}
K.~Q. Weinberger, J.~Blitzer, and L.~K. Saul.
\newblock Distance metric learning for large margin nearest neighbor
  classification.
\newblock In {\em Advances in neural information processing systems}, pages
  1473--1480, 2005.

\bibitem{xie_fusing_2010}
S.~Xie, S.~G. Shan, X.~L. Chen, and J.~Chen.
\newblock Fusing local patterns of gabor magnitude and phase for face
  recognition.
\newblock {\em {IEEE} Transactions on Image Processing}, 19(5):1349--1361,
  2010.

\bibitem{yi_learning_2014}
D.~Yi, Z.~Lei, S.~Liao, and S.~Z. Li.
\newblock Learning face representation from scratch.
\newblock {\em arXiv preprint arXiv:1411.7923}, 2014.

\bibitem{zhang_histogram_2007}
B.~C. Zhang, S.~G. Shan, X.~L. Chen, and W.~Gao.
\newblock Histogram of {G}abor phase patterns (hgpp): a novel object
  representation approach for face recognition.
\newblock {\em {IEEE} Transactions on Image Processing}, 16(1):57--68, 2007.

\bibitem{zhao_face_2003}
W.~Y. Zhao, R.~Chellappa, P.~J. Phillips, and A.~Rosenfeld.
\newblock Face recognition: A literature survey.
\newblock {\em ACM Computing Surveys}, 35(4):399--458, 2003.

\end{thebibliography}
}

\end{document}